\documentclass{article}
\usepackage{ijcai25}

\usepackage{times}
\usepackage{soul}
\usepackage{url}
\usepackage[hidelinks]{hyperref}
\usepackage[utf8]{inputenc}
\usepackage[small]{caption}
\usepackage{graphicx}
\usepackage{amsmath}
\usepackage{amsthm}
\usepackage{booktabs}
\usepackage{algorithm}
\usepackage{algorithmic}
\usepackage[switch]{lineno}
\usepackage{multirow}
\usepackage{tabularx}
\usepackage{xcolor}
\usepackage{textcomp}

\title{MADP: Multi-Agent Deductive Planning for Enhanced Cognitive-Behavioral Mental Health Question Answer}

\author{
    Qi Chen$^*$
    \and
    Dexi Liu$^*$\\
    \affiliations
    Jiangxi University of Finance and Economics\\
    \emails
    chenqi.1997@qq.com, dexi.liu@163.com
}

\begin{document}

\maketitle

\begin{abstract}
 The Mental Health Question Answer (MHQA) task requires the seeker and supporter to complete the support process in one-turn dialogue. Given the richness of help-seeker posts, supporters must thoroughly understand the content and provide logical, comprehensive, and well-structured responses. Previous works in MHQA mostly focus on single-agent approaches based on the cognitive element of Cognitive Behavioral Therapy (CBT), but they overlook the interactions among various CBT elements, such as emotion and cognition. This limitation hinders the models' ability to thoroughly understand the distress of help-seekers. To address this, we propose a framework named Multi-Agent Deductive Planning (MADP), which is based on the interactions between the various psychological elements of CBT. This method guides Large Language Models (LLMs) to achieve a deeper understanding of the seeker's context and provide more personalized assistance based on individual circumstances. Furthermore, we construct a new dataset based on the MADP framework and use it to fine-tune LLMs, resulting in a specialized model named MADP-LLM. We conduct extensive experiments, including comparisons with multiple LLMs, human evaluations, and automatic evaluations, to validate the effectiveness of the MADP framework and MADP-LLM.

\end{abstract}

\definecolor{mypink}{RGB}{250,219,223}
\definecolor{myblue}{RGB}{223,236,247} 
\definecolor{mygreen}{RGB}{226,240,219} 

\section{Introduction}
\begin{figure}[t]
\centering
\centerline{\includegraphics[scale=0.65]{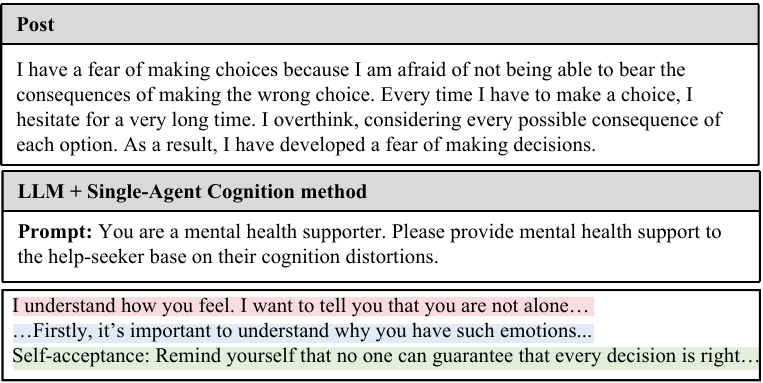}}
\caption{The diagram illustrates the limitations of single-agent methods in mental health support.}
\label{ov}
\end{figure}

The Mental Health Question Answer (MHQA) task is an interactive approach where help-seekers share psychological struggles, and supporters respond as needed \cite{bib18}. This method does not require real-time interactions, making it flexible, thereby enabling its widespread use on social forums such as Reachout, Reddit, and Yixinli. Typically, MHQA aims to complete the entire support process within one-turn dialogue. Given the rich content of help-seeking posts, supporters must thoroughly understand the content and integrate multiple strategies to provide logical and comprehensive responses. Therefore, building a MHQA system requires not only strong comprehension capabilities but also robust logical narrative skills.

Recent studies have shown that large language models (LLMs) excel in semantic understanding and text generation \cite{bib6,bib34}. In the latest research on mental health support \cite{bib34,bib51}, a novel approach has been introduced that integrates the cognitive elements of Cognitive Behavioral Therapy (CBT) into the prompts of a single agent to generate responses with better analysis and guidance. This generation process is illustrated in Figure 1. Although these approachs can provide improved analysis and guidance by leveraging cognitive elements and LLMs' semantic understanding, it still faces limitations. Specifically, the sections on:

1) \textbf{The emotional resonance (Pink)} is not specific enough. For example, the response ``I understand how you feel'' does not clarify what exactly is being understood. Help-seekers who seek emotional support usually expect empathetic support, such as active listening, genuine understanding, and comforting responses, for example, ``I can understand your fear of making choices.''

2) \textbf{The interpretation resonance (Blue)} is insufficient. For instance, the statement ``it’s important to understand why you have such emotions'' identifies a general issue but fails to address the help-seeker's specific emotional struggles. A deeper analysis, such as ``This unease about decision outcomes often leads to avoidance, as a way to escape potential failure or disappointment,'' can better help help-seekers confront cognitive distortions.

3) \textbf{The Guidance resonance (Green)} is not sufficiently targeted. For example, the advice ``Remind yourself that no one can guarantee that every decision is right'' is too general. In contrast, targeted guidance, such as ``You can start by practicing decision-making with small, everyday matters. Even if the outcome is perceived as `wrong,' the impact will be minimal, helping you gradually build confidence,'' can help help-seekers accept correct cognitive perspectives.

These limitations arise because help-seeking posts often contain extensive background information, personal experiences, and emotional expressions, which require a more comprehensive analysis considering the interactions among multiple CBT elements in the help-seeking posts.

\begin{figure}[t]
\centering
\centerline{\includegraphics[scale=0.40]{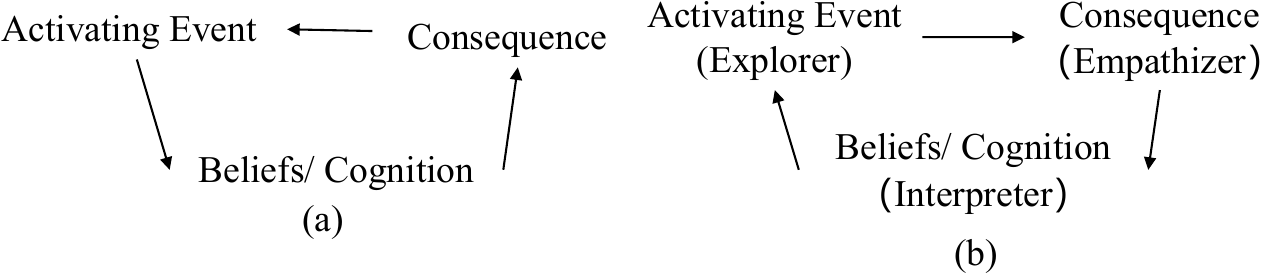}}
\caption{(a) depicts the traditional \textbf{\underline{ABC}} model in CBT, showing the sequential relationship between activating events \textbf{\underline{A}}, beliefs/cognitions \textbf{\underline{B}}, and consequences \textbf{\underline{C}}. (b) presents the proposed multi-agent framework, illustrating the interactions among the three agents (arrows indicate the flow of information).}
\label{ov}
\end{figure}

To address these limitations, we adopt the ABC model from CBT, which emphasizes the interaction between activating events \textbf{\underline{A}}, beliefs/cognitions \textbf{\underline{B}}, and consequences \textbf{\underline{C}}. Traditionally, the ABC model posits that patients' emotions and behaviors are formed in the sequence of \textbf{\underline{A \textrightarrow B \textrightarrow C \textrightarrow A}}, as illustrated in Figure 2(a). Based on this, we propose a reverse analysis approach, analyzing the formation mechanism of patients' psychological states through the process of \textbf{\underline{A \textrightarrow C \textrightarrow B \textrightarrow A}}, as illustrated in Figure 2(b).
To operationalize this approach, we introduce three specialized agents within the MADP framework: \textbf{Explorer ($A_{EX}$)}, corresponding to \textbf{\underline{A}}, identifies external events that trigger emotional reactions; \textbf{Empathizer ($A_{EM}$)}, corresponding to \textbf{\underline{C}}, uncovers and responds to the seeker's emotional consequences; and \textbf{Interpreter ($A_{IN}$)}, corresponding to \textbf{\underline{B}}, interprets cognitive patterns and provides insights to address distortions. Together, these agents enable a comprehensive understanding of the seeker's psychological state and facilitate targeted interventions. Additionally, to transform the multi-agent dialogue deductive into well-structured responses, we introduce a support planning method. Therefore, the multi-agent dialogue deduction and support planning constitute the MADP framework.

In addition to the MADP framework, we aim to create a cost-effective, open-source MHQA model for local deployment. While closed-source LLMs like GPT4o offer high-quality support, their large parameter sizes and commercial constraints restrict local deployment. Open-source LLMs with smaller parameter sizes (e.g., LLaMA3-8B, GLM4-9B) can be fine-tuned on help-seeking posts, but their responses lack structured guidance. To address this, we use the MADP framework commercial LLMs to generate high-quality support responses and create a mental health support dataset. This dataset is then used to fine-tune smaller open-source LLMs, enhancing their ability to provide emotional support.

The main contributions of this paper are: 1)We propose the Multi-Agent Deductive Planning (MADP) framework, a CBT-based multi-agent reasoning strategy that analyzes interactions among multiple CBT elements to comprehensively understand and respond to help-seeking posts. 2) We construct a high-quality mental health dataset and fine-tune MADP-LLM, enhancing emotional reasoning capabilities while reducing deployment costs. 3) Extensive experiments demonstrate the effectiveness of MADP in providing personalized, empathetic, and targeted mental health support.

\section{Related Work}
\subsection{Advances in Mental Health Support Research}
Early research on generative mental health support primarily focused on mental health dialogue \cite{bib8} , using predefined support strategies to simulate the support process and alleviate the speaker's negative emotions. Subsequent research has focused on developing technologies to better understand the needs of help-seekers \cite{bib47,bib49}. Among these efforts, Peng et al. \shortcite{bib47} proposed a global-to-local hierarchical graph network to better understand the global causes and local intentions of help-seekers, enhancing the effectiveness of emotional support. Tu et al. \shortcite{bib49} employed a commonsense reasoning model, COMET, for fine-grained emotion understanding and developed a hybrid response strategy integrating emotional information. These methods significantly improve dialogue generation by deepening the understanding of help-seekers' emotions and intentions. However, these approaches have notable limitations. Specifically, they face two major challenges: First, their generalization capabilities are limited due to heavy reliance on predefined model frameworks and datasets, which restricts their adaptability to diverse and dynamic scenarios. Second, existing research primarily focuses on surface-level emotions, often neglecting users' deeper psychological needs and complex emotional states.

The potential of LLMs in mental health support has gradually emerged, demonstrating their ability to provide scalable and accessible assistance \cite{bib41,bib42}. In mental health dialogue research, Zhang et al. \shortcite{bib50} proposed the ESCoT scheme, creating the first mental health dialogue dataset with chain-of-thought annotations to improve interpretability. Xiao et al. \shortcite{bib51} developed the HealMe model, which guides clients through cognitive restructuring for self-exploration. In MHQA research, Wang et al. \shortcite{bib6} proposed the Cue-CoT method, using a single agent to infer help-seekers' states and generate responses. Additionally, Na et al. \shortcite{bib34} integrated a CBT-based single-agent approach into prompts to improve LLM response effectiveness. However, these single-agent methods primarily focus on the cognitive element of CBT, neglecting interactions between other CBT elements. Moreover, their emphasis on rapid response generation results in limited analytical and empathetic score \cite{bib5}.

\subsection{Potential and Challenges of Multi-Agent Systems}

The collaborative potential of LLMs has led to significant advancements in multi-agent systems, such as improved task quality, role-playing frameworks, and debate mechanisms \cite{bib14,bib21,bib22,bib52,bib53,bib24} . However, current systems often fail to strategically manage dialogue progression or address complex needs in MHQA, such as deeply exploring user contexts and expressing empathy \cite{bib58,bib33}.

\begin{figure}[t]
\centering
\centerline{\includegraphics[scale=0.40]{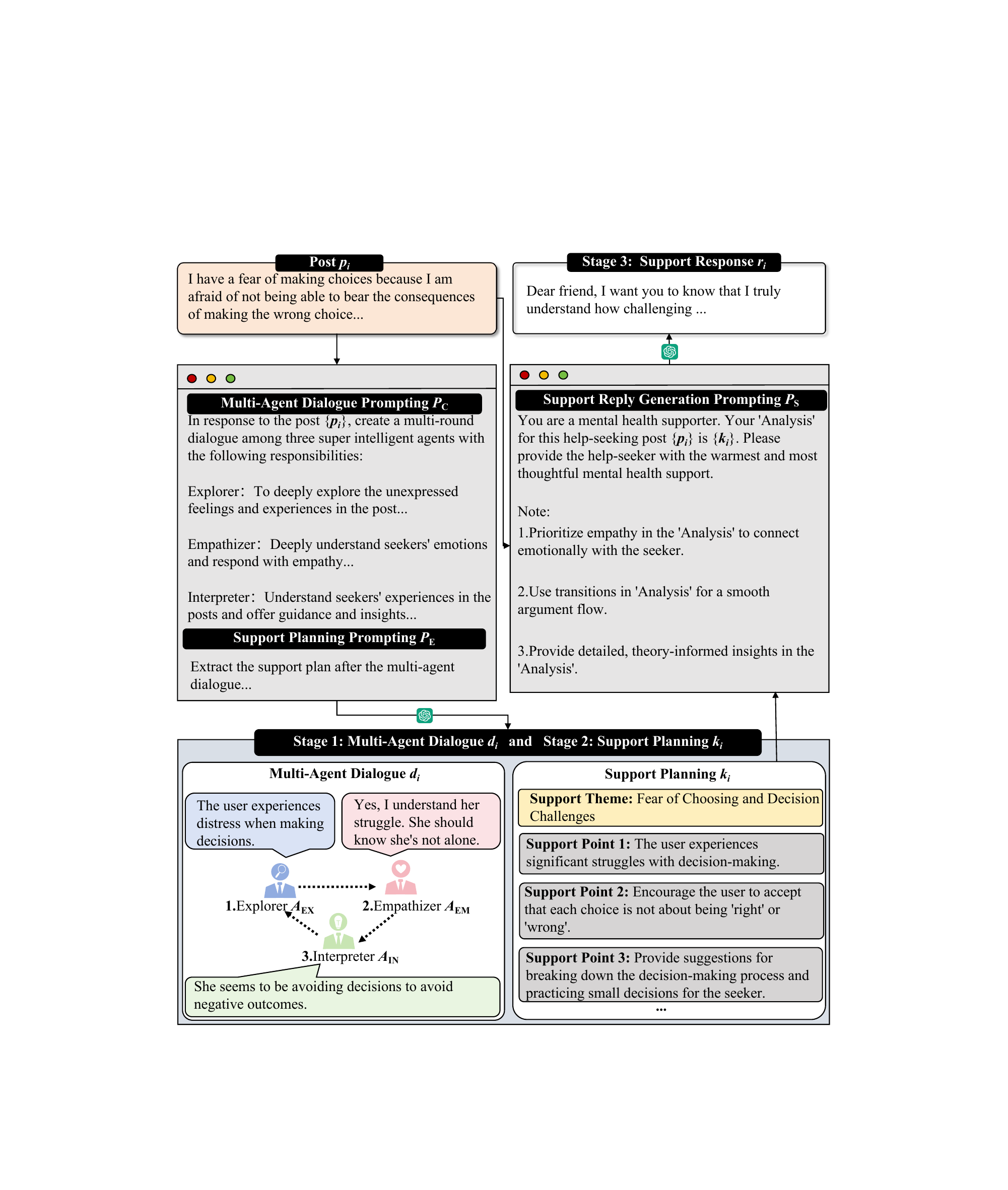}}
\caption{An example of prompt for the MADP Framework.}
\label{ov}
\end{figure}

\section{MADP Framework}
This paper introduces the MADP framework for MHQA, illustrated in the upper portion of Figure 3. The framework consists of three stages: multi-agent dialogue deduction, support planning, and support generation. During the dialogue deduction phase, multiple agents with diverse expertise analyze the issues presented in the help-seeking post. Their goal is to deeply understand the psychological challenges, provide specific empathy, and offer targeted guidance. The support planning stage employs prompts to direct LLMs in deriving a mental health support plan from these dialogues. Finally, the support generation stage leverages the help-seeking post and the support plan to generate a response.

\subsection{Multi-Agent Dialogue Deduction}

In this phase, three independent agents are tasked with analyzing the user's problem from their respective perspectives via iterative dialogue. The design of the agents is informed by two key theories: the ABC model from CBT, which emphasizes the interactions among activating events (\textbf{\underline{A}}), beliefs/cognitions (\textbf{\underline{B}}), and consequences (\textbf{\underline{C}}), and Sharma et al.\shortcite{bib7}'s communication mechanisms theory. The details are as follows:

\textbf{Explorer ($A_{EX}$):}  
Corresponding to \textbf{\underline{A}} in the ABC model, $A_{EX}$ explores the external activating events that trigger the user's emotional reactions, including feelings and experiences that may not have been explicitly mentioned. Grounded in the ABC theory, $A_{EX}$ focuses on identifying the external events (\textbf{\underline{A}}) that elicit the seeker's emotional responses. At the same time, $A_{EX}$ adheres to the ``Explorations'' mechanism from the communication framework \cite{bib7}. This mechanism involves delving into unstated feelings and experiences in the seeker's post.

\textbf{Empathizer ($A_{EM}$):}  
Corresponding to \textbf{\underline{C}} in the ABC model, $A_{EM}$ uncovers the seeker's emotional consequences and expresses understanding and empathy for their emotional state. Based on the ABC theory, $A_{EM}$ focuses on the emotional outcomes (\textbf{\underline{C}}) of the seeker's experiences. It also follows the ``Emotional Reactions'' mechanism from the communication framework \cite{bib7}. This mechanism involves conveying the supporter's emotional experience after reading the seeker's post to establish an emotional connection. Notably, $A_{EM}$'s primary goal is to build trust and a sense of security through emotional bonding, rather than providing specific explanations or advice. However, $A_{EM}$ may occasionally offer emotional insights to strengthen the connection with the seeker.

\textbf{Interpreter ($A_{IN}$):}  
Corresponding to \textbf{\underline{B}} in the ABC model, $A_{IN}$ interprets the feelings, experiences, and emotions uncovered by $A_{EX}$ and $A_{EM}$ from a cognitive perspective. Based on the ABC theory, $A_{IN}$ identifies the seeker's cognitive distortions (\textbf{\underline{B}}), provides insights to help them confront these distortions, and assists in adopting healthier perspectives \cite{bib45}. Additionally, $A_{IN}$ follows the ``Interpretations'' mechanism from the communication framework \cite{bib7}. This mechanism involves conveying an understanding of the seeker's feelings and experiences through their post.

Under the guidance of the Multi-Agent Dialogue Deduction prompt $P_\text{C}$, the multi-agent system \( A = \{A_{\text{EX}}, A_{\text{EM}}, A_{\text{IN}}\} \) generates multi-agent dialogue content \( d_i \) based on the help-seeking post \( p_i \). The objective can be viewed as follows:

\begin{equation}
f_{\text{LLM}}(p_i, P_\text{C}) \rightarrow d_i \tag{1}
\end{equation}

\subsection{Support Planning}
In the planning phase, the MADP framework avoids immediately generating support responses. Instead, it processes the multi-agent dialogue from Phase 1 to construct a support plan, guiding the LLM to produce a logically clear, rich, and complete response, as illustrated in Figure 3.

At this stage, based on the help-seeking post $p_i$ and the dialogue content $d_i$, the support plan $k_i$ is generated by the LLM under the guidance of the Support Planning Prompt $P_\text{E}$, as follows:
\begin{equation}
f_{\text{LLM}}(p_i, d_i, P_\text{E}) \rightarrow k_i \tag{2}
\end{equation}

\subsection{Support Response Generation}

The task of the Support Response Generation phase is to generate a complete mental health support response $r_i$ by the LLM based on the help-seeking post $p_i$ and the support plan $k_i$, guided by the Support Generation Prompt $P_\text{S}$, as follows:
\begin{equation}
f_{\text{LLM}}(p_i, k_i, P_\text{S}) \rightarrow r_i \tag{3}
\end{equation}

\section{MADP-LLM}
We propose using closed-source LLMs guided by the MADP framework as ``teacher models'' to construct a mental health support dataset with planning capabilities. This dataset will then be used to fine-tune small open-source LLMs to improve their performance in mental health support, as illustrated in Figure 4.

\begin{figure}[t]
\centering
\centerline{\includegraphics[scale=0.38]{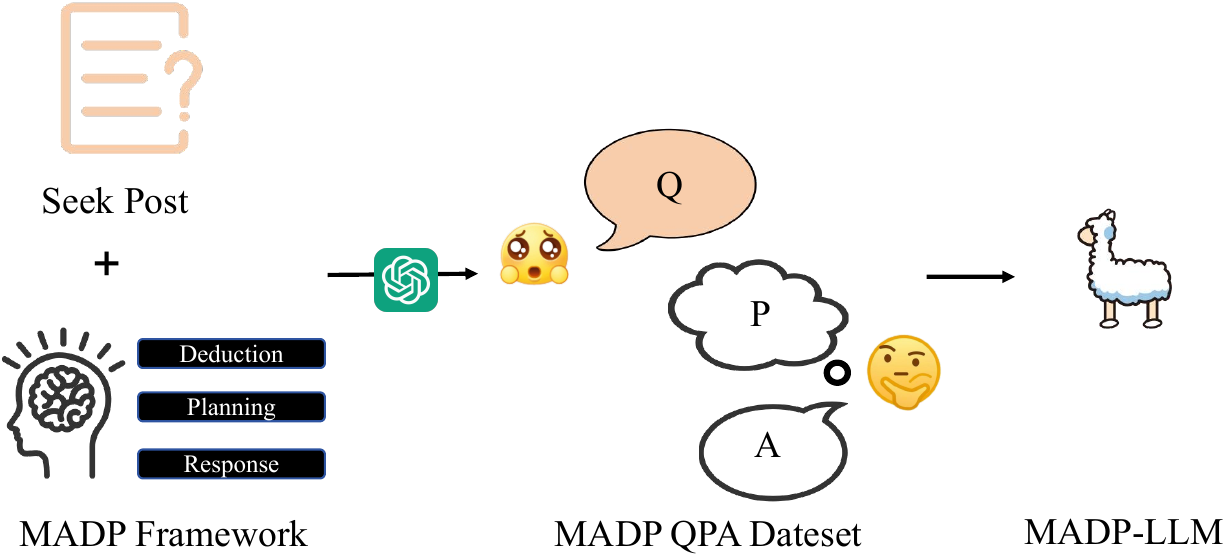}}
\caption{An overview of training MADP-LLM. It first utilizes help-seeking posts and MADP Prompt to generate MADP planning and answers, and then fine-tuning MADP-LLM.}
\label{ov}
\end{figure}

\subsection{MADP Dataset}

The capabilities of LLMs in language generation and reasoning provide a practical approach to generating high-quality MHQA data \cite{bib15,bib20,bib50}. However, existing MHQA datasets primarily consist of help-seeking posts and response posts, and they lack support plans that guide models to generate high-quality mental health support responses \cite{bib7,bib18}.

As shown in the lower half of Figure 4, this paper leverages the MADP framework to generate a MHQA dataset with support plans, defined as 
$D_{MADP}=\{ (p_i; k_i; v_i) \}$
, in which each element $(p_i; k_i; v_i)$ represents the i-th post $p_i$, its corresponding support plan $k_i$, and support response $v_i$.

\subsection{MADP-LLM}

The basic architecture of the LLMs employed in this study is based on the Transformer-Decoder autoregressive framework, which predicts the next word in a sequence. The MADP dataset is used as the fine-tuning dataset, with the fine-tuning instruction  $Z= [\text{Plan first}; \text{them respond}], X=(p_i; Z; k_i; v_i) $. The prediction process can be formulated as follows:

\begin{equation}
Pr(X) = \prod_{n=1}^{N} Pr(x_n | x_0, x_1, \dots, x_{n-1}) \tag{4}
\end{equation}

Here, $N$ represents the total length of $X$, and $x_n$ is the $n$-th token in $X$. This paper integrates instruction fine-tuning \cite{bib35} and LoRA fine-tuning techniques \cite{bib36}. Instruction fine-tuning provides explicit task instructions during training to align model outputs with task requirements. LoRA enhances performance by adding trainable low-rank matrices $A$ and $B$ to each layer, adjusting the output while keeping the pre-trained weight matrix $W$ unchanged. The parameter update $\Delta W$ is expressed as:

\begin{equation}
\Delta W = A \times B \tag{5}
\end{equation}

Additionally, for training MADP-LLM, this paper utilizes the cross-entropy loss. The loss can be computed using the following equation:

\begin{equation}
Loss = -\sum_{n=1}^{N} \log Pr(x_n | x_0, x_1, \dots, x_{n-1}) \tag{6}
\end{equation}

\begin{table*}[ht]
\centering
\tabcolsep=0.05cm
\renewcommand{\arraystretch}{1.2} %
\caption{Zero-Shot Evaluation Results of MADP Framework (“-MADP” indicates the use of the MADP framework; bold font represents the best results).}

\scalebox{0.8}{
\begin{tabular}{llccccc}
\hline
\textbf{Dataset} & \textbf{Model} & \textbf{Analytical} & \textbf{Empathy} & \textbf{Guidance} & \textbf{Comprehensive} & \textbf{Average} \\
\hline
\multirow{6}{*}{ EMH Dataset} & GPT4o & 7.50 & 8.05 & 7.32 & 7.70 & 7.64 \\
 & GPT4o-MADP & \textbf{7.90 (+5.33\%)} & \textbf{8.52 (+5.84\%)} & \textbf{7.54 (+3.00\%)} & \textbf{8.06 (+4.68\%)} & \textbf{8.01 (+4.74\%)} \\
 & LLaMA3-8b & 6.95 & 7.89 & 6.84 & 7.30 & 7.25 \\
 & LLaMA3-8b-MADP & 7.37 (+6.04\%) & 8.12 (+2.92\%) & 7.17 (+4.82\%) & 7.64 (+4.66\%) & 7.57 (+4.55\%) \\
 & GLM4-9b & 6.91 & 7.54 & 6.79 & 7.13 & 7.09 \\
 & GLM4-9b-MADP & 7.19 (+4.05\%) & 7.75 (+2.79\%) & 7.10 (+4.57\%) & 7.44 (+4.35\%) & 7.37 (+3.91\%) \\
\hline
 & \textbf{Average Improvement} & (+5.14\%) & (+3.85\%) & (+4.13\%) & (+4.56\%) & (+4.42\%) \\
\hline
\multirow{6}{*}{PsyQA Dataset} & GPT4o & 7.55 & 8.11 & 7.42 & 7.78 & 7.71 \\
 & GPT4o-MADP & \textbf{7.95 (+6.00\%)} & \textbf{8.34 (+3.60\%)} & \textbf{7.60 (+3.83\%)} & \textbf{8.01 (+4.03\%)} & \textbf{7.97 (+4.35\%)} \\
 & LLaMA3-8b & 6.97 & 7.60 & 6.83 & 7.18 & 7.14 \\
 & LLaMA3-8b-MADP & 7.34 (+5.61\%) & 8.08 (+2.41\%) & 7.08 (+3.51\%) & 7.55 (+3.42\%) & 7.51 (+3.69\%) \\
 & GLM4-9b & 7.27 & 7.74 & 7.28 & 7.49 & 7.45 \\
 & GLM4-9b-MADP & 7.41 (+7.24\%) & 7.82 (+3.71\%) & 7.29 (+7.36\%) & 7.57 (+6.17\%) & 7.52 (+6.06\%) \\
\hline
 & \textbf{Average Improvement} & (+6.28\%) & (+3.24\%) & (+4.90\%) & (+4.54\%) & (+4.74\%) \\
\hline
\textbf{Average Improvement (Overall)} & & (+5.71\%) & (+3.54\%) & (+4.52\%) & (+4.55\%) & (+4.58\%) \\
\hline
\end{tabular}}
\end{table*}

\section{Experimental}

\subsection{Datasets}
We evaluate our approach using the following datasets:

\textbf{EMH Dataset}: Collected from 55 mental health-related subreddits\cite{bib7}, it contains 9,243 help-seeking post-response pairs annotated with three communication types (Explorations, Emotional Reactions, Interpretations) and three intensity levels (None, Weak, Strong). For testing, we randomly select 10 pairs from each level and type, yielding a total of 90 pairs.

\textbf{PsyQA Dataset}: Collected from the Yixinli platform. \cite{bib18}, it includes 22,346 samples classified into 9 categories. We randomly select 10 samples per category. For each help-seeking post that contains multiple answers, we manually select the best response as the ideal answer, resulting in a test set of 90 question-answer pairs.

\textbf{PsyQA-MADP and EMH-MADP Datasets}: Constructed using the MADP framework, these datasets extend PsyQA and EMH with generated support plans and responses. For each dataset, we randomly select 50 posts per category (9 categories in total), generate responses, and divide them into 80\% training and 20\% testing sets, yielding 360 training pairs and 90 testing pairs.

\subsection{Evaluation Methods}
\textbf{Automatic Evaluations.} we utilize LLMs as automatic evaluators, as traditional metrics like BLEU and ROUGE exhibit limited alignment with human judgments \cite{bib6,bib31,bib32,bib33}. We evaluate four key dimensions of the generated responses on a scale from 1 to 10 with a score interval of 0.1: (1) \textbf{Analytical} (depth of understanding personalized issues); (2) \textbf{Empathy} (level of human care and sincerity); (3) \textbf{Guidance} (quality of personalized and professional advice); (4) \textbf{Comprehensive} (overall performance, including coherence and fluency).

\textbf{Human Evaluations.} We invited three evaluators with training in psychological assessment to conduct the evaluation. Eighteen help-seeking posts were randomly selected from the EMH and PsyQA test sets for comparison. To ensure objectivity, the evaluators conducted the assessments blindly, unaware of the sources of the responses.

\subsection{Baseline Models and Comparison Frameworks}
We assess the MADP framework with GPT4o, LLaMA3-8b, and GLM4-9b as baseline architectures, using Claude 3.5 Sonnet as the evaluation model. LLaMA3-8b is an open-source LLM primarily trained on English data, with multilingual capabilities. GLM4-9b, another open-source LLM, is mainly trained on Chinese data and also supports multiple languages. For MADP-LLM, GPT4o is used to generate the MADP dataset, while LLaMA3-8b and GLM4-9b are fine-tuned as smaller-scale open-source LLMs.

To highlight the benefits of the MADP framework, we compare it with the following baseline methods:

\textbf{M-Cue CoT Prompt Framework} \cite{bib6}: This framework first deduces the psychological state of the help-seeker and then produces a supportive response based on the inferred state.

\textbf{CBT Prompt Framework} \cite{bib34}: This framework utilizes principles from CBT to generate supportive responses.

\textbf{Standard Prompt}: A single-agent prompt developed based on the ``Psychologist'' and ``Mental Health Adviser'' templates from the awesome-chatgpt-prompts platform. The prompt is ``You are a mental health supporter, please give the seeker mental health support in the most warm and caring way.'' This method serves as the baseline in subsequent comparative analyses.

\subsection{Training Configuration}
The MADP-LLM model is built using the LLaMA Factory. The hyperparameters are set as follows: learning rate = $5 \times 10^{-5}$, temperature = 1.0, batch size = 1, and epoch = 5. Training is performed on a Linux system with an Nvidia GeForce RTX 3090 GPU (24GB memory).

\begin{table*}[ht]
\tabcolsep=0.05cm
\renewcommand{\arraystretch}{1.2} %
\centering
\caption{Comparison of Different Mental Health Support Frameworks (“-MADP” denotes the use of the MADP framework, “-M-Cue” denotes the use of the M-Cue CoT Prompt framework, “-CBT” denotes the use of the CBT Prompt framework, and “Human Response” refers to responses sourced from the original dataset. Bold font highlights the best results.)}
\scalebox{0.8}{
\begin{tabular}{llccccc}
\hline

\textbf{Dataset} & \textbf{Model} & \textbf{Analytical} & \textbf{Empathy} & \textbf{Guidance} & \textbf{Comprehensive} & \textbf{Average} \\
\hline
\multirow{5}{*}{EMH Dataset} & GPT4o & 7.50 & 8.05 & 7.32 & 7.70 & 7.64 \\
 & GPT4o-M-Cue & 7.54 (+0.53\%) & 8.28 (+2.86\%) & 7.12 (-2.73\%) & 7.71 (+0.13\%) & 7.66 (+0.26\%) \\
 & GPT4o-CBT & 7.63 (+1.73\%) & 8.07 (+0.25\%) & 7.29 (+0.41\%) & 7.74 (+0.52\%) & 7.68 (+0.52\%) \\
 & \textbf{GPT4o-MADP} & \textbf{7.90 (+5.33\%)} & \textbf{8.52 (+5.84\%)} & \textbf{7.54 (+3.01\%)} & \textbf{8.06 (+4.68\%)} & \textbf{8.01 (+4.84\%)} \\
 & Human Response & 4.90 & 5.02 & 3.70 & 4.46 & 4.52 \\
\hline
\multirow{5}{*}{PsyQA Dataset} & GPT4o & 7.55 & 8.11 & 7.42 & 7.78 & 7.71 \\
 & GPT4o-M-Cue & 7.78 (+3.05\%) & 7.90 (-2.59\%) & 7.57 (+2.02\%) & 7.81 (+0.39\%) & 7.76 (+0.65\%) \\
 & GPT4o-CBT & 7.89 (+4.50\%) & 7.88 (-2.84\%) & 7.59 (+2.29\%) & 7.84 (+0.77\%) & 7.80 (+1.10\%) \\
 & \textbf{GPT4o-MADP} & \textbf{7.95 (+5.30\%)} & \textbf{8.34 (+2.84\%)} & \textbf{7.60 (+2.43\%)} & \textbf{8.01 (+2.96\%)} & \textbf{7.97 (+3.37\%)} \\
 & Human Response & 6.90 & 6.88 & 6.61 & 6.78 & 6.79 \\
\hline
\end{tabular}}
\end{table*}

\subsection{Comparison with Different LLM Backbones}
As shown in Table 1, the MADP framework significantly enhances LLM performance across multiple dimensions in MHQA tasks. Applied to GPT4o, LLaMA3-8b, and GLM4-9b, MADP improved Analytical, Empathy, Guidance, and Comprehensive score  by 5.71\%, 3.54\%, 4.52\%, and 4.55\% on average, respectively. This demonstrates MADP's compatibility and adaptability across LLMs and languages. MADP's multi-agent dialogue simulation enhances complex information processing, particularly in mental health scenarios, enabling accurate issue identification and deeper psychological understanding. Specialized agents like the ``Empathizer'' and ``Interpreter'' further align responses with users' emotional needs.

\textbf{MADP bridges language-specific performance gaps and activates cross-language potential.} While LLaMA3-8b outperformed GLM4-9b on the English EMH dataset, GLM4-9b led on the Chinese PsyQA dataset, highlighting the importance of language-specific training. MADP significantly improved LLMs' performance on non-primary languages. For instance, GLM4-9b-MADP surpassed LLaMA3-8b on the English dataset in Analytical, Guidance, and Comprehensive score (by 3.45\%, 3.80\%, and 1.91\%, respectively). Similarly, LLaMA3-8b-MADP outperformed GLM4-9b on the Chinese dataset in Analytical, Empathy, and Comprehensive score (by 0.96\%, 4.39\%, and 0.80\%, respectively). This demonstrates MADP's ability to unlock LLMs' multilingual potential.

\textbf{A performance gap persists between small-scale and large-scale LLMs.} Despite MADP's improvements, models like LLaMA3-8b and GLM4-9b still lag behind GPT4o, likely due to differences in capacity, training data, and architecture. This gap highlights the need for further research, motivating the design of MADP-LLM to enhance smaller models' capabilities.

\subsection{Comparing with Other Prompt Methods}

To assess the effectiveness of the MADP framework, we compared it with several mental health support frameworks using GPT4o, the best-performing model. The comparison results, shown in Table 2, demonstrate that GPT4o-MADP significantly surpasses other frameworks in analytical ability, empathy ability, guidance ability, and comprehensive ability. The detailed analysis is as follows:

\textbf{CBT significantly improves LLMs' support responses.} Both MADP and CBT prompt frameworks \cite{bib34} are based on CBT theory. GPT4o-CBT improved Analytical, Guidance, and Comprehensive score by 1.73\%, 0.41\%, and 0.52\% on the EMH dataset, and by 4.50\%, 2.29\%, and 0.77\% on the PsyQA dataset, demonstrating CBT's effectiveness in problem analysis and guidance.

\textbf{CBT prompts do not fully stimulate LLMs' empathy.} GPT4o-CBT improved empathy by only 0.25\% on the EMH dataset but decreased by 2.84\% on the PsyQA dataset. This is because CBT frameworks \cite{bib34} overlook the interactions between key elements, such as emotion and cognition. This limitation hinders the models' ability to thoroughly understand the distress of help-seekers. In contrast, GPT4o-MADP improved empathy by 5.84\% and 3.60\% on the EMH and PsyQA datasets, respectively, through its empathizer agent.

\textbf{GPT4o-MADP demonstrates stability and adaptability across Datasets, while GPT4o-M-Cue and GPT4o-CBT struggle with variability.} The English EMH dataset features direct and colloquial posts, posing challenges for models to deeply empathize and understand. On the EMH dataset, GPT4o-M-Cue's Guidance score decreased by 2.73\%, as it prioritizes emotional states over actual needs. Meanwhile, GPT4o-CBT's Empathy score improved by only 0.25\%, showing limited emotional resonance despite some guidance. In contrast, the Chinese PsyQA dataset features complex posts with nuanced emotional expressions and rich cultural contexts, requiring models to capture both surface meanings and underlying emotions and intentions. On the PsyQA dataset, GPT4o-M-Cue's Empathy score decreased by 2.59\%, reflecting its inability to adapt to unique emotional expressions and cultural details. Similarly, GPT4o-CBT's Empathy score decreased by 2.84\%, as its single-agent structure struggles with complex scenarios and multi-layered emotions. In contrast, GPT4o-MADP demonstrates stability and adaptability on both datasets, leveraging its multi-agent structure to comprehensively understand user needs and provide personalized support.

\textbf{Human responses underperform compared to LLM-based methods.} This stems from supporters' lack of professional training and standardized counseling skills\cite{bib34}, highlighting a shortage of trained professionals. The growing demand for online support underscores the need for advanced MHQA models with robust analytical, empathetic, and guidance capabilities.

\begin{table*}[ht]
\centering
\caption{Experimental Results After Fine-Tuning on Different Datasets. (The subscript ``ft'' denotes a fine-tuned model. ``-MAr'' indicates that the training data for fine-tuning comprises only responses generated by the MADP framework. ``-MApr'' indicates that the training data incorporates both responses generated by the MADP framework and support planning. ``-H'' denotes that the training data comprises the original human responses from the dataset. LLMs with the suffix ``-MApr'' and ``-MAr'' are designated as MADP-LLMs. The values in parentheses indicate the improvement relative to the corresponding LLMs using the single-agent prompt method, e.g., LLaMA3-8b and GLM4-9b.)}
\scalebox{0.8}{
\begin{tabular}{llccccc}
\hline
\textbf{Dataset} & \textbf{Model} & \textbf{Analytical} & \textbf{Empathy} & \textbf{Guidance} & \textbf{Comprehensive} & \textbf{Average} \\
\hline
\multirow{8}{*}{EMH Dataset} & GPT4o & 7.50 & 8.05 & 7.32 & 7.70 & 7.64 \\
 & LLaMA3-8b & 6.95 & 7.89 & 6.84 & 7.30 & 7.25 \\
 & LLaMA3-8bft-MAr & 7.51 (+8.06\%) & 8.26 (+4.69\%) & 7.09 (+3.65\%) & 7.69 (+5.34\%) & 7.64 (+5.42\%) \\
 & LLaMA3-8bft-MApr & \textbf{7.60 (+9.35\%)} & \textbf{8.30 (+5.20\%)} & 7.20 (+5.26\%) & 7.77 (+6.44\%) & 7.72 (+6.52\%) \\
 & LLaMA3-8bft-H & 4.72 & 4.09 & 3.52 & 4.04 & 4.09 \\
 & GLM4-9b & 6.91 & 7.54 & 6.79 & 7.13 & 7.09 \\
 & GLM4-9bft-MAr & 7.53 (+8.97\%) & 8.28 (+9.81\%) & 7.19 (+5.89\%) & 7.73 (+8.42\%) & 7.68 (+8.32\%) \\
 & GLM4-9bft-MApr & 7.59 (+9.84\%) & \textbf{8.30 (+10.08\%)} & \textbf{7.40 (+8.98\%)} & \textbf{7.85 (+10.10\%)} & \textbf{7.79 (+9.76\%)} \\
 & GLM4-9bft-H & 4.50 & 4.27 & 3.40 & 3.97 & 4.04 \\
\hline
\multirow{8}{*}{PsyQA Dataset} & GPT4o & 7.55 & 8.11 & 7.42 & 7.78 & 7.72 \\
 & LLaMA3-8b & 6.97 & 7.60 & 6.83 & 7.18 & 7.14 \\
 & LLaMA3-8bft-MAr & 7.44 (+6.74\%) & 7.95 (+4.61\%) & 7.08 (+3.66\%) & 7.54 (+5.01\%) & 7.50 (+5.00\%) \\
 & LLaMA3-8bft-MApr & \textbf{7.81 (+12.05\%)} & \textbf{8.22 (+8.16\%)} & \textbf{7.37 (+7.91\%)} & \textbf{7.86 (+9.47\%)} & \textbf{7.82 (+9.38\%)} \\
 & LLaMA3-8bft-H & 5.98 & 5.82 & 5.63 & 5.79 & 5.81 \\
 & GLM4-9b & 7.27 & 7.74 & 7.28 & 7.49 & 7.45 \\
 & GLM4-9bft-MAr & 7.47 (+2.75\%) & 8.03 (+3.75\%) & 7.18 (-1.37\%) & 7.64 (+2.00\%) & 7.58 (+1.81\%) \\
 & GLM4-9bft-MApr & 7.59 (+4.40\%) & 8.08 (+4.39\%) & 7.30 (+0.27\%) & 7.71 (+2.94\%) & 7.67 (+3.02\%) \\
 & GLM4-9bft-H & 6.53 & 6.47 & 6.24 & 6.41 & 6.41 \\
\hline
\end{tabular}}
\end{table*}
\begin{table*}[h]
\centering
\caption{Manual Evaluation Results (with "-MADP" indicating the use of the MADP framework; bold font indicates optimal results)}
\scalebox{0.8}{
\begin{tabular}{llccccc}
\hline
\textbf{Dataset} & \textbf{Model} & \textbf{Analytical} & \textbf{Empathy} & \textbf{Guidance} & \textbf{Comprehensive} & \textbf{Average} \\
\hline
\multirow{2}{*}{EMH Dataset} & GPT4o & 7.36 & 7.56 & 7.51 & 7.60 & 7.51 \\
 & \textbf{GPT4o-MADP} & \textbf{7.85 (+6.66\%)} & \textbf{8.12 (+7.41\%)} & \textbf{7.74 (+3.06\%)} & \textbf{7.99 (+5.13\%)} & \textbf{7.93 (+5.56\%)} \\
\hline
\multirow{2}{*}{PsyQA Dataset} & GPT4o & 7.11 & 7.31 & 7.27 & 7.33 & 7.25 \\
 & \textbf{GPT4o-MADP} & \textbf{7.85 (+10.41\%)} & \textbf{8.04 (+9.99\%)} & \textbf{7.68 (+5.64\%)} & \textbf{7.85 (+7.09\%)} & \textbf{7.86 (+8.27\%)} \\
\hline
\end{tabular}}
\end{table*}

\subsection{MADP-LLM Evaluation}

The MADP Datasets significantly improves the mental health support capabilities of LLMs when used for fine-tuning. As shown in Table 3, training solely with MADP-generated replies improves model performance, and incorporating both MADP-generated support plans and replies during training results in even greater improvements. This demonstrates that the combination of both components is essential for maximizing the effectiveness of LLMs in mental health support tasks.

\textbf{Training data with support planning significantly improves LLMs' abilities.} Models trained with support planning (marked as -MApr) significantly surpassed those trained without support planning (marked as -MAr or unmarked) across various ability metrics. For example, smaller models like LLaMA3-8bft-MApr and GLM4-9bft-MApr matched or even exceeded the performance of larger models like GPT4o. 

\textbf{Training data with support planning makes LLMs' performance more stable and effective in psychological support.} On the PsyQA dataset, the guidance ability score of GLM4-9b before fine-tuning was 7.28, while GLM4-9bft-MAr (trained with only responses) achieved a score of 7.18. This suggests that using only responses as training data may result in adaptation challenges. However, when support planning was added, GLM4-9bft-MApr's guidance ability increased to 7.30. This demonstrates that combining support planning with responses improves the adaptability of LLMs, making their performance more stable and effective.

\textbf{Models trained with human responses performed poorly.} This underscores the critical importance of dataset quality in training effectiveness. Currently, high-quality MHQA datasets are limited. By introducing the MADP framework to generate high-quality training data, this study seeks to enhance the performance of LLMs in mental health support tasks, mitigating the scarcity of existing resources.

\textbf{The MADP Datasets significantly enhances LLM performance in cross-lingual MHQA tasks.} On the English dataset, Chinese LLMs (GLM4-9bft-MApr and GLM4-9bft-MAr) showed improvements of 9.76\% and 8.32\% in Average score, outperforming English LLMs (LLaMA3-8bft-MApr and LLaMA3-8bft-MAr) with improvements of 6.52\% and 5.42\%. On the Chinese dataset, English LLMs improved by 9.38\% and 5.00\%, surpassing Chinese LLMs in Average score, which improved by 3.02\% and 1.81\%. This highlights the effectiveness of cross-lingual training.

\subsection{Human Evaluation}
The results in Table 4 demonstrate that, by leveraging the interactions among various CBT elements, the responses from MADP are more preferred by human judges compared to those from baseline models. In particular, compared to baseline, our model has shown significant improvements in all aspects, especially on Analytical and Empathy metrics. This indicates that our framework is better at understanding issues and providing comfort to others.

\section{Conclusion}
In this paper, we propose the Multi-Agent Deductive Planning (MADP) framework to improve MHQA systems by addressing the limitations of analytical ability. Inspired by the ABC model from Cognitive Behavioral Therapy, the MADP framework employs three specialized agents: Explorer, Empathizer, and Interpreter. These agents interact to simulate the reverse process of \textbf{\underline{A \textrightarrow B \textrightarrow C \textrightarrow A}}, facilitating a deeper understanding of the help-seeker's emotional and cognitive states. We leverage GPT4o to construct a reasoning chain dataset for supervised training. Based on this dataset, we fine-tune a compact model, MADP-LLM, to facilitate cost-effective deployment. Experiments validate the effectiveness of MADP, demonstrating an average performance improvement of 5\% across key score. Our work contributes to personalized, accessible, and high-quality mental health support grounded in psychological principles.

\section*{Ethical Statement}
The datasets used in this paper are publicly available, with sensitive personal information removed during creation. This work is strictly for academic research, exploring the potential of LLMs in providing mental health support. However, we caution against direct real-world application due to ethical risks, including the possibility of inappropriate content from baseline models and the limitations of LLMs in replacing professional human judgment. The generated responses should always be reviewed or modified by humans to ensure quality and reliability.

\bibliographystyle{named}
\bibliography{ijcai25}

\end{document}